\documentclass[letterpaper, 10 pt, conference]{ieeeconf}  
\IEEEoverridecommandlockouts
\usepackage{cite}
\usepackage{amsmath,amssymb,amsfonts,hyperref}
\usepackage{algorithmic}
\usepackage{graphicx}
\usepackage{textcomp}
\usepackage{xcolor}
\usepackage{colortbl}
 \usepackage{multirow} 
 \usepackage{colortbl} 
\usepackage[normalem]{ulem}
 \useunder{\uline}{\ul}{}
\def\BibTeX{{\rm B\kern-.05em{\sc i\kern-.025em b}\kern-.08em
    T\kern-.1667em\lower.7ex\hbox{E}\kern-.125emX}}
\begin{document}
\author{Fei Yan$^{*1}$, Yuhong He$^{*2}$, Keyu Chen†$^{1}$, En Cheng$^{1}$ and Jikang Ma$^{1}$
\thanks{This work is supported in part by the National Natural Science Foundation of China (62271425 and 62071402). *euual contribution. †Corresponding author.}
\thanks{$^{1}$School of Informatics, Xiamen University, Xiamen, China
        }%
\thanks{$^{2}$School of Computer Science and Engineering, Northeastern University, Shenyang, China}
}
\title{ \LARGE \bf
Adaptive Frequency Enhancement Network for Single Image Deraining
}

\maketitle

\begin{abstract}


Image deraining aims to improve the visibility of images damaged by rainy conditions, targeting the removal of degradation elements such as rain streaks, raindrops, and rain accumulation. While numerous single image deraining methods have shown promising results in image enhancement within the spatial domain, real-world rain degradation often causes uneven damage across an image's entire frequency spectrum, posing challenges for these methods in enhancing different frequency components. In this paper, we introduce a novel end-to-end Adaptive Frequency Enhancement Network (AFENet) specifically for single image deraining that adaptively enhances images across various frequencies. We employ convolutions of different scales to adaptively decompose image frequency bands, introduce a feature enhancement module to boost the features of different frequency components and present a novel interaction module for interchanging and merging information from various frequency branches. Simultaneously, we propose a feature aggregation module that efficiently and adaptively fuses features from different frequency bands, facilitating enhancements across the entire frequency spectrum. This approach empowers the deraining network to eliminate diverse and complex rainy patterns and to reconstruct image details accurately. Extensive experiments on both real and synthetic scenes demonstrate that our method not only achieves visually appealing enhancement results but also surpasses existing methods in performance. The source code is available at \href{https://github.com/yanfefei/FENet} {https://github.com/yanfefei/AFENet}.
\end{abstract}

\section{Introduction}
\begin{figure}[t]
	
	\begin{minipage}{0.32\linewidth}
		\vspace{3pt}
		\centerline{\includegraphics[width=\textwidth]{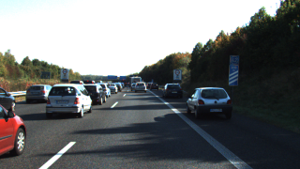}}
		\centerline{(a) GT}
	\end{minipage}
	\begin{minipage}{0.32\linewidth}
		\vspace{3pt}
		\centerline{\includegraphics[width=\textwidth]{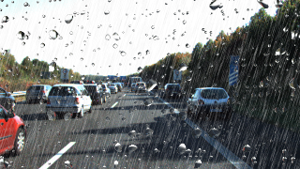}}
	 
		\centerline{(b) Rainy}
	\end{minipage}
	\begin{minipage}{0.32\linewidth}
		\vspace{3pt}
		\centerline{\includegraphics[width=\textwidth]{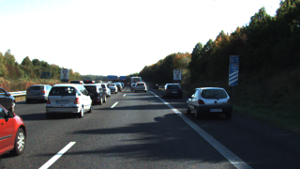}}
	 
		\centerline{(c) Deraining}
	\end{minipage}
 
	\caption{Example of rain removal in rainy conditions.  }
	\label{fig4}
\end{figure}

Single image deraining focuses on reconstructing a rain-free image from a degraded rainy image captured in rainy conditions, as shown in Fig. \ref{fig4}. Since both the clear image and the nature of rain degradation are typically unknown, image deraining presents as an ill-posed and challenging problem \cite{yang2020single}. Rain degradation significantly hinders computer vision tasks such as image classification, object detection, and semantic segmentation \cite{testolina2023selma,10400436}. For example, vision-based intelligent driving systems are severely affected in rainy conditions, making it imperative to remove rain and reconstruct high-quality, rain-free images effectively. Traditional methods often rely on mathematical statistics to derive diverse priors by exploring the physical characteristics of rain streaks and accumulation. Various priors, including layer priors with the Gaussian mixture model (GMM) \cite{Gauss}, discriminative sparse coding (DSC) \cite{sparsity}, and high-frequency priors \cite{frequency}, have been proposed to regularize and separate rain streaks. However, these traditional approaches struggle with complex rainy scenes in real-world scenarios.

In recent times, numerous deep learning-based methods have been developed to learn the transformation from a rainy to a clear image in a data-driven manner \cite{chen2022snowformer, chen2023sparse, peng, engin2018cycle, zhang2020deblurring, ancuti2020ntire, yan2023textual, li2023ntire,he2022deep}. These methods have significantly advanced the removal of rain streaks and image enhancement in the spatial domain. Yet, the distribution of rain artifacts varies across different frequency ranges, posing a challenge to these methods in effectively enhancing different frequency components within the spatial domain. In real-world rainy scenarios, rain streaks, raindrops, and rain accumulation coexist \cite{yang2020single,li2021comprehensive}, with rain streaks predominantly present in high-frequency image areas, whereas rain accumulation and raindrops are more prevalent in mid and low frequencies.

\begin{figure}[t]
\centerline{\includegraphics[width=0.5\textwidth,height=0.15\textheight]{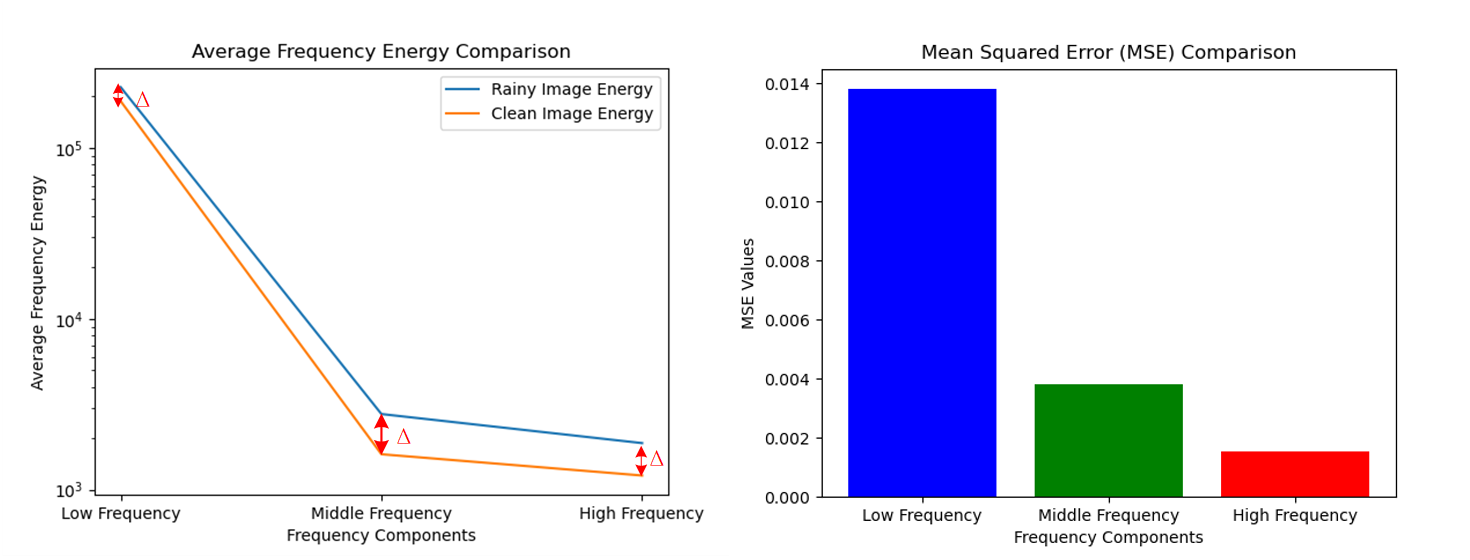}}
\caption{Frequency Analysis of Various Rainy Images (including rain streaks, raindrops, and rain accumulation) and Clean Images: The left side of the analysis highlights the differences in energy distribution among various frequency components. On the right side, the focus is on the variations in Mean Squared Error (MSE) between rainy and clean images across various frequencies. Here, $\Delta$ signifies the absolute difference in rainy degradation across these different frequency components.}
\label{fig1}
\end{figure}

To elucidate this, we conduct a frequency analysis in various rainy scenes, analyzing over 1000 pairs of rainy and derained images including conditions with rain streaks, raindrops, and rain accumulation. We utilized the Discrete Cosine Transform (DCT) to separate the frequencies into high, middle, and low bands, calculating the frequency energy and energy error between paired rainy and clean images across different frequencies. As illustrated in Fig. \ref{fig1}, there are notable energy discrepancies at different frequencies. This significant variance in errors across frequency bands highlights the need for targeted enhancement in each band. Moreover, since rain streaks and background details are often intertwined in the spatial domain, existing methods struggle to completely eliminate rain streaks and recover structural information in complex rainy scenarios \cite{li2021comprehensive,yang2020single,su2023survey}. Most previous works have focused predominantly on spatial information, overlooking distinct frequency information. The differences between rainy and clear image pairs are more distinctly defined in the frequency domain. Consequently, a novel image deraining network that can adaptively enhance the frequency becomes essential for achieving high-quality image reconstruction.

To fulfill this objective, we introduce a novel Adaptive Frequency Enhancement Network (AFENet) for single image deraining, specifically designed to enhance features across different frequency components. Our approach begins with the use of scale-variable convolutions to decompose the image into three distinct frequency components. We then present a novel Feature Enhancement Module (FEM), which effectively segments and manages information within different frequency domain branches, thereby enhancing features across these components. Subsequently, we propose an innovative Feature Aggregation Module (FAM) that facilitates the interactive fusion of features from various frequency branches. This comprehensive enhancement across all frequency components enables the deraining network to effectively remove a wide array of complex rainy patterns and reconstruct image details. Extensive testing on both real and synthetic scenes has shown that our method not only achieves visually appealing results but also surpasses existing methods in effectiveness.

In summary, the key contributions of our research are as follows:

\begin{itemize}
\item We introduce the Adaptive Frequency Enhancement Network (AFENet), a novel approach designed to effectively enhance features in different frequency components, thus aiding in reconstructing high-quality, clean images.

\item A novel Frequency Enhancement Module and Feature Aggregation Module are developed, aiming at proficient feature enhancement and fusion across various frequency branches.

\item Through comprehensive experiments, we demonstrate that our method not only achieves aesthetically pleasing enhancement results but also consistently outperforms existing techniques in both real and synthetic scenes.
\end{itemize}

\section{Related Work}

Over the past decade, significant advancements have been achieved in the field of image restoration under adverse weather conditions  \cite{yan2023textual,peng2021ensemble,wang2022self,wang2022dense, wang2023decoupling, peng2024lightweight,peng2024towards,peng2024efficient,he2024latent,jiang2024dalpsr,wu2024mixnet,zhang2023federated,wang2023decoupling,yi2023frequency,yi2021structure,yi2021efficient}, with significant advancements in single image deraining methods garnering widespread attention and achieving notable success \cite{chen2022snowformer, engin2018cycle, zhang2020deblurring, ancuti2020ntire, wang2023brightness}. Current single image deraining techniques can be broadly classified into traditional and deep learning-based methods. Traditional methods focus on the statistical analysis of rain streaks and background scenes, while deep learning-based methods leverage data-driven approaches that utilize deep neural networks to autonomously extract hierarchical features and develop more sophisticated models to restore rainy images into clean ones.

\subsection{Traditional Single Image Deraining}
Traditional deraining approaches often involve utilizing diverse priors to create additional constraints, forming a cost function, and optimizing it. Luo \textit{et al.} \cite{luo2015removing} proposed a single-image deraining algorithm based on a non-linear generative model, using dictionary learning and discriminative codes to represent rain layers sparsely. Li \textit{et al.} \cite{Gauss} tackled rain streak removal as a layer decomposition problem, employing Gaussian mixture models as a layer prior. Zhu \textit{et al.} \cite{sparsity} decomposed rainy images into a rain-free background and a rain-streak layer. These methods, while effective on synthetic datasets, often face challenges in real-world scenarios due to mixed and non-uniform degradations and the lack of sufficient paired data for training.

\subsection{Deep Learning-based Single Image Deraining}
Recently, numerous deep learning methods have emerged, demonstrating exceptional performance \cite{semi-derain, URML, RESCAN, PreNet, sparsity, MSPFN, MPRNet,yi2021structure,yi2021efficient}. Yang \textit{et al.} \cite{wang2017deep} developed a network for joint rain detection and removal, handling heavy rain and accumulation. Fu \textit{et al.} \cite{DerainNet} introduced a deep detail network for rain streak removal, focusing on high-frequency details. Zhang \textit{et al.} \cite{DIDMDN} designed a multi-stream dense network for density-aware rain streak removal. Jiang \textit{et al.} \cite{MSPFN} proposed a multi-scale progressive fusion strategy to capture global textures of rain streaks. In \cite{MPRNet}, a multi-stage architecture was introduced for contextualized feature learning. Ye \textit{et al.} \cite{ye2021closing} employed a bidirectional disentangled translation network for joint rain generation and removal. Xiao \textit{et al.} \cite{IDT} proposed a Transformer-based image de-raining architecture. Chen \textit{et al.} \cite{chen2023sparse} introduced a Sparse Sampling Transformer for adaptive degradation sampling. Although these methods perform well in spatial domain detail reconstruction, they face challenges in fully restoring images in all frequency bands under complex, real-world rainy conditions, resulting in rain residual and detail distortion in complex and real-world rainy conditions.

\begin{figure*}[t]
\centering
\includegraphics[width=0.95\textwidth]{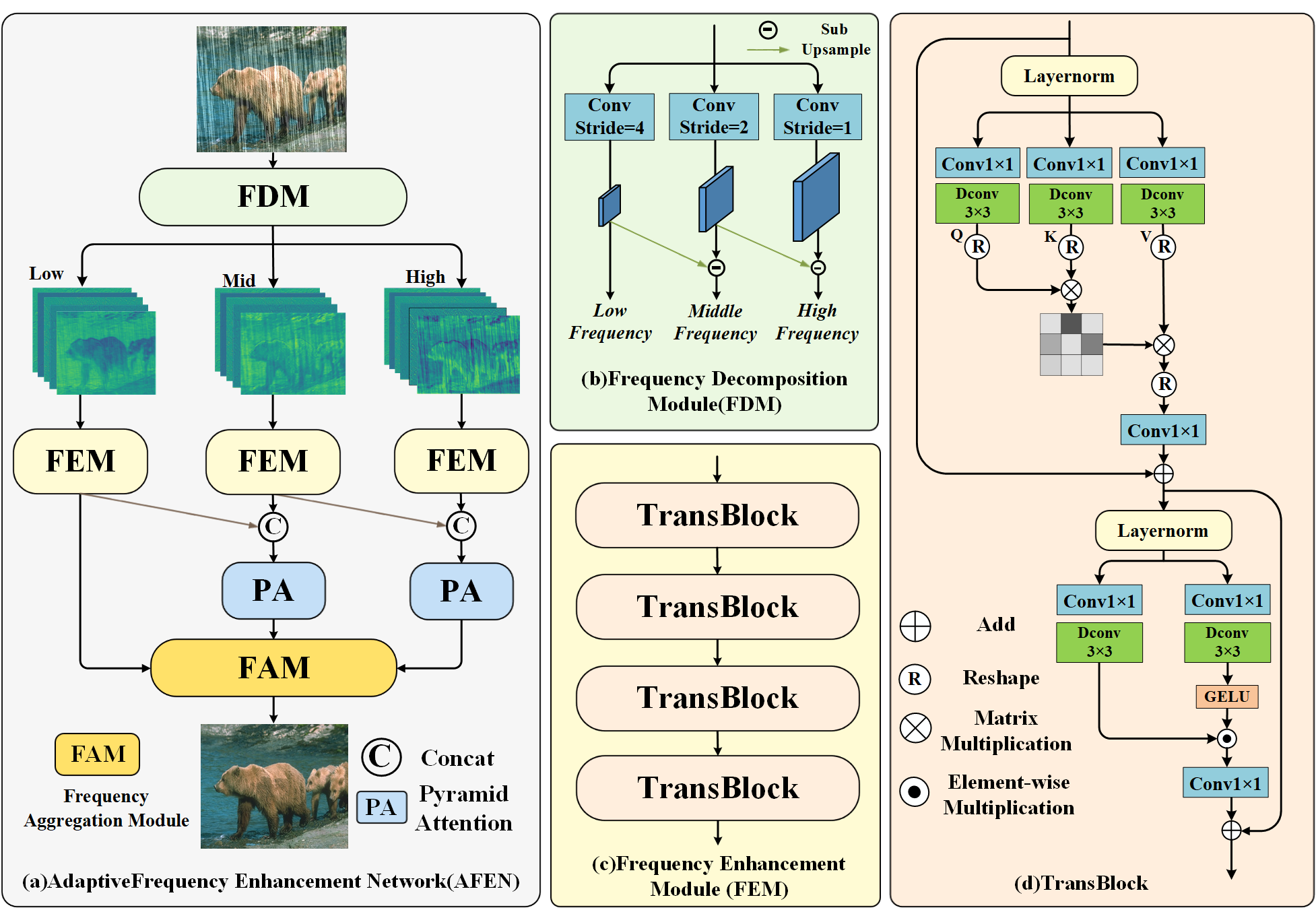}
\centering
\caption{The architecture of our proposed Adaptive Frequency Enhancement Network (AFENet) includes three key modules: the Frequency Decomposition Module (FDM), which employs a multi-branch architecture to decompose the rainy image content in the frequency domain, thereby enhancing texture details across all frequency components; the Frequency Enhancement Module (FEM), designed to improve the feature representation capabilities; and the Frequency Aggregation Module (FAM), which aggregates the enhanced frequency components from various scales of the rainy image. }
\label{main figure}
\end{figure*}

\section{Proposed Method}

In this section, we present an in-depth introduction to the newly proposed Adaptive Frequency Enhancement Network (AFENet). The comprehensive architecture of AFENet is illustrated in Fig. \ref{main figure}. AFENet is fundamentally composed of three key components: the Frequency Decomposition Module (FDM), the Frequency Enhancement Module (FEM), and the Frequency Aggregation Module (FAM).

\subsection{Frequency Decomposition Module}
To further elaborate on the innovative design of the Frequency Decomposition Module (FDM), as depicted in Fig. \ref{main figure}(b), we have developed it with the aim of achieving an exhaustive representation across all frequency bands. FDM ingeniously segments rainy images into three distinct frequency bands: high, middle, and low. Diverging from conventional image processing methods like the Discrete Cosine Transform (DCT) \cite{zhang2023multi} or the wavelet transform \cite{nakamura2020time}, our approach opts for convolutions with varied receptive fields to adaptively discern features across different frequency ranges. This choice is driven by a critical insight: traditional methods, while precise, often lead to a loss of essential information imperative for low-level restoration tasks due to their deterministic mathematical framework and limited applicability to specific tasks. This is further examined in our ablation studies detailed in Section \ref{Ablation Study}. The unique stratification afforded by FDM paves the way for creating specialized frequency-enhancement branches, each tailored to learn representations pertinent to their designated frequency band.

Furthermore, to safeguard pivotal details, we have integrated a novel implicit downsampling method for decomposing frequency-specific features. In this nuanced approach, we utilize convolutional layers with incrementally larger strides for extracting low-frequency features, effectively isolating the high-frequency elements by eliminating lower-frequency components from the initial features. This process entails employing a convolutional layer with a stride of 4 for the extraction of the low-frequency component, denoted as $f_l$. The mid-frequency component $f_m$ is then isolated by subtracting $f_l$ from the original features, correspondingly downsampled with a stride of 2. In a similar way, the high-frequency component $f_h$ is derived by subtracting the features downsampled with a stride of 2 from the original rainy image's unsampled features, thereby preserving the original spatial dimensions. The formulation of FDM is as follows:

\begin{equation}
\begin{aligned}
& f_l=\operatorname{Conv} \downarrow_2\left(\operatorname{Conv} \downarrow_2(I)\right), \\
& f_m=\operatorname{Conv} \downarrow_2(I)-\operatorname{Conv} \downarrow_2\left(\operatorname{Conv} \downarrow_2(I)\right) \uparrow_2, \\
& f_h=\operatorname{Conv}(I)-\operatorname{Conv} \downarrow_2(I) \uparrow_2,
\end{aligned}
\end{equation}
$\operatorname{Conv}\downarrow_2$ represents the convolution layer with a stride of 2, and $\operatorname{Conv}$ signifies a convolution layer without downsampling. $\uparrow_2$ denotes the bilinear upsampling operation. This frequency decomposition approach offers two primary benefits. Firstly, unlike traditional multi-branch networks, it improves network efficiency by reducing feature resolution. Secondly, it boosts performance by embracing multi-scale representation learning, thereby effectively tackling degradations across different frequencies \cite{pan2022fast}.

\subsection{Frequency Enhancement Module}
Continuing from the previous explanation, the real-world application of image deraining, as demonstrated in Fig. \ref{fig1}, clearly highlights the necessity of addressing a full-spectrum degradation problem. It becomes imperative to enhance features across the entire frequency spectrum to effectively tackle this issue. Therefore, in our proposed architecture, as shown in Fig. \ref{main figure}(c), the Frequency Enhancement Module (FEM) is utilized to augment each set of features corresponding to high, mid, and low frequencies.

Recognizing that rainy images often suffer from global degradation, our Feature Extraction Module (FEM) is specifically designed to extract both global and local features, an approach that addresses the complex nature of image de-raining. The FEM employs an innovative transformer block named the Multi-Dconv Head Transposed Attention (MDTA), which marks a significant advancement over traditional CNNs. The core strength of the MDTA lies in its ability to handle global features more effectively than CNNs, primarily due to the transformer architecture's inherent prowess in this aspect.

MDTA stands out by utilizing self-attention (SA) mechanisms across channels as opposed to spatial dimensions, enabling the computation of cross-covariance. This results in the generation of an attention map that implicitly captures the global context of the image. A key feature of MDTA is the integration of depth-wise convolutions that focus on the local context. These convolutions precede the computation of feature covariance, which is crucial for constructing the global attention map. The depth-wise convolutions thus ensure a balanced focus on both local and global aspects of the image features, enhancing the module's effectiveness in de-raining applications. The operational mechanism of MDTA is defined as follows:
\begin{equation}
\begin{aligned}
& \hat{{X}}=W_p \operatorname{Attention}(\hat{{Q}}, \hat{{K}}, \hat{{V}})+{X} \\
& \operatorname{Attention}(\hat{{Q}}, \hat{{K}}, \hat{{V}})=\hat{{V}} \cdot \operatorname{Softmax}(\hat{{K}} \cdot \hat{{Q}} / \alpha)
\end{aligned}
\end{equation}
where $X$ and $\hat{X}$ are the input and output feature maps. $\hat{{Q}} \in {R}^{\hat{H} \hat{W} \times \hat{C}}$; $\hat{K} \in {R}^{\hat{C}  \times \hat{H} \hat{W} }$, ${{X}} \in {R}^{\hat{H} \times \hat{W} \times \hat{C}}$ and $\hat{{V}} \in {R}^{\hat{H} \hat{W} \times \hat{C}}$ matrices are obtained after reshaping tensors from the original size ${R}^{\hat{H} \times \hat{W} \times \hat{C}}$. Here, $\alpha$ is a learnable scaling parameter to adjust the magnitude of the dot product of ${K}$ and ${Q}$ before applying the softmax function.

Furthermore, to capture local features, the FEM introduces Gated-Dconv Feed-Forward Network (GDFN) \cite{restormer}. The GDFN captures information flow across different hierarchical levels within our framework. This method enables each level to concentrate on intricate details, complementing other levels and thereby enriching the features with contextual information. GDFN is formulated as:
\begin{align}\label{eqn:int1}
\hat{{X}} & =W_p^0 \text { Gating }({X})+{X} \\
\operatorname{Gating}({X}) & =\phi\left(W_d^1 W_p^1({L N}({X}))\right) \odot W_d^2 W_p^2({L N}({X}))\nonumber
\end{align}
where $\odot$ denotes element-wise multiplication, $\phi$ represents the GELU non-linearity, and LN is the layer normalization. enables each branch to selectively fuse the corresponding frequency components at various stages, significantly enhancing the representation capability of each branch.

\begin{figure}[t]
\centerline{\includegraphics[width=\columnwidth]{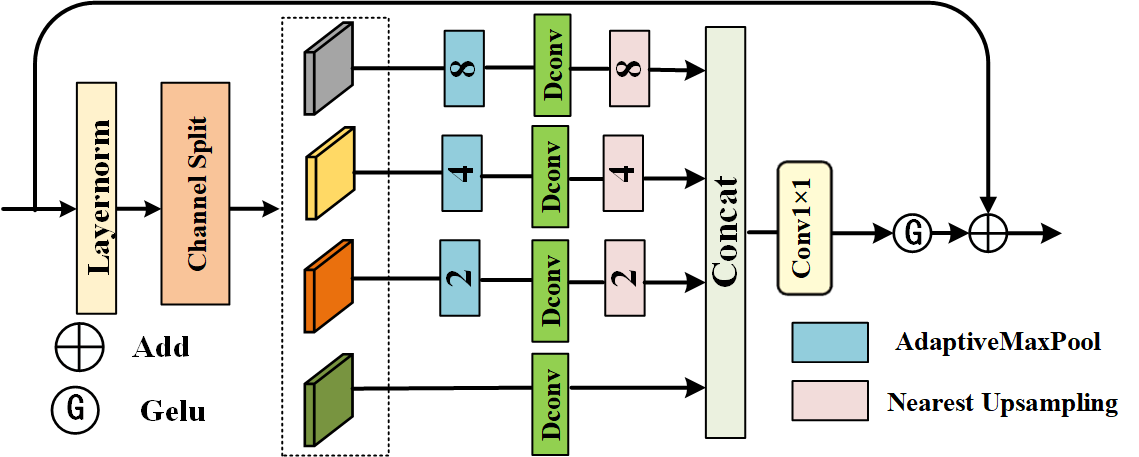}}
\caption{Frequency Aggregation Module (FAM): a multi-scale feature generation unit to produce features at multiple scales.}
\label{fre agg}
\end{figure}

Moreover, given the inherent challenge of restoring high-frequency feature components, we introduce a strategy to enhance these complex frequency components from coarse to fine. Specifically, we facilitate interaction between different frequency components, progressively leveraging the enhanced lower-frequency representations to enhance the higher-frequency components. This is achieved by concatenating these frequency components together. Subsequently, we employ Pyramid Attention to refine the details in the higher-frequency components further. This strategy not only addresses the difficulty of restoring high-frequency details but also ensures a comprehensive and integrated enhancement across all frequency bands. We define this enhancement and interaction process as follows:
\begin{equation}
\begin{aligned}
&\hat{{F_l}} =\operatorname{enhance}({f_l}),\\
&\hat{{F_m}} =\operatorname{PA}
(\operatorname{enhance}({F_m})\odot \operatorname{enhance}({F_l})),\\
&\hat{{F_h}} =\operatorname{PA}
(\operatorname{enhance}({F_h})\odot \operatorname{enhance}({F_m})) 
\end{aligned}
\end{equation}
where enhance(·) denotes a set of several frequency enhancement modules (FEM), and PA(·) denotes Pyramid Attention, which can be found in Fig. \ref{main figure}(a).

\subsection{Frequency Aggregation Module}

In addressing the varied extent of frequency information loss in different areas of rainy images, our method effectively utilizes long-range dependencies derived from multi-scale feature representations. This methodology significantly improves the identification and restoration of crucial features for reconstructing a clean image. This paper provides an in-depth exploration of the Frequency Aggregation Module (FAM), as depicted in Fig. \ref{fre agg}. To find an optimal balance between model complexity and the achievement of a pyramid-like feature representation, we initially implemented a channel-wise split operation on the normalized input features. Subsequently, a multi-scale feature generation unit is utilized to produce features at various scales. Specifically, a 3 × 3 depth-wise convolution is applied to the first segment. The remaining segments undergo a series of upsampling and downsampling operations, each followed by a depth-wise convolution.

\begin{table*}[t]
\centering
\caption{Quantitative comparisons with existing state-of-the-art deraining methods. Average refers to the average performance of the five benchmark datasets. The \textbf{bold} and \underline{underline} represent the best and second-best performance respectively. }
\label{quantitative}  
\small
\centering
\begin{tabular}{ccccccccccc|cc}
\hline
                & \multicolumn{2}{c}{Test100}                           & \multicolumn{2}{c}{Rain100H}                          & \multicolumn{2}{c}{Rain100L}                          & \multicolumn{2}{c}{Test1200}                          & \multicolumn{2}{c|}{Test2800}                          & \multicolumn{2}{c}{\cellcolor[HTML]{EFEFEF}Average}           \\
Metrics         & PSNR                      & SSIM                      & PSNR                      & SSIM                      & PSNR                      & SSIM                      & PSNR                      & SSIM                      & PSNR                      & SSIM                       & \cellcolor[HTML]{EFEFEF}PSNR  & \cellcolor[HTML]{EFEFEF}SSIM  \\ \hline
DerainNet \cite{DerainNet}       & 22.77                     & 0.810                     & 14.92                     & 0.592                     & 27.03                     & 0.884                     & 23.38                     & 0.835                     & 24.31                     & 0.861                      & \cellcolor[HTML]{EFEFEF}22.48 & \cellcolor[HTML]{EFEFEF}0.796 \\
SEMI \cite{SEMI}           & 22.35                     & 0.788                     & 16.56                     & 0.486                     & 25.03                     & 0.842                     & 26.05                     & 0.822                     & 24.43                     & 0.782                      & \cellcolor[HTML]{EFEFEF}22.88 & \cellcolor[HTML]{EFEFEF}0.744 \\
DIDMDN \cite{DIDMDN}          & 22.56                     & 0.818                     & 17.35                     & 0.524                     & 25.23                     & 0.741                     & 29.95                     & 0.901                     & 28.13                     & 0.867                      & \cellcolor[HTML]{EFEFEF}24.64 & \cellcolor[HTML]{EFEFEF}0.770 \\
UMRL \cite{URML}           & 24.41                     & 0.829                     & 26.01                     & 0.832                     & 29.18                     & 0.923                     & 30.55                     & 0.910                     & 29.97                     & 0.905                      & \cellcolor[HTML]{EFEFEF}28.02 & \cellcolor[HTML]{EFEFEF}0.880 \\
RESCAN \cite{RESCAN}          & 25.00                     & 0.835                     & 26.36                     & 0.786                     & 29.80                     & 0.881                     & 30.51                     & 0.882                     & 31.29                     & 0.904                      & \cellcolor[HTML]{EFEFEF}28.59 & \cellcolor[HTML]{EFEFEF}0.858 \\
SPANet \cite{SPANet}          & 23.17                     & 0.833                     & 26.54                     & 0.843                     & 32.20                     & 0.951                     & 31.36                     & 0.912                     & 30.05                     & 0.922                      & \cellcolor[HTML]{EFEFEF}28.66 & \cellcolor[HTML]{EFEFEF}0.892 \\
PReNet \cite{PreNet}          & 24.81                     & 0.851                     & 26.77                     & 0.858                     & 32.44                     & 0.950                     & 31.36                     & 0.911                     & 31.75                     & 0.916                      & \cellcolor[HTML]{EFEFEF}29.43 & \cellcolor[HTML]{EFEFEF}0.897 \\
MSPFN \cite{MSPFN}           & 27.50                     & 0.876                     & 28.66                     & 0.860                     & 32.40                     & 0.933                     & 32.39                     & 0.916                     & 32.82                     & 0.930                      & \cellcolor[HTML]{EFEFEF}30.75 & \cellcolor[HTML]{EFEFEF}0.903 \\
CRDNet \cite{peng}          & 27.72 & 0.887 & 28.63 & 0.872 & 33.28 & 0.952 & 32.44 & 0.913 & 33.39 & 0.935 & \cellcolor[HTML]{EFEFEF}31.09     & \cellcolor[HTML]{EFEFEF}0.912     \\
MPRNet \cite{MPRNet}          & {30.27}                     & 0.897                     & {30.41}                     & 0.890                     & 36.40                     & 0.965                     & {32.91}                     & {0.916}                     & {33.64}                     & {0.938}                      & \cellcolor[HTML]{EFEFEF}32.73
& \cellcolor[HTML]{EFEFEF}0.921 \\
IDLIR \cite{IDLIR}           & 28.33                     & 0.894                     & 29.33                     & 0.886                     & 35.72                     & 0.965                     & 32.06                     &  \underline{0.917}                     & 32.93                     & 0.936                      & \cellcolor[HTML]{EFEFEF}31.67 & \cellcolor[HTML]{EFEFEF}0.920 \\
Uformer-B \cite{uformer} & 29.90                     & {0.906}                     & 30.31                     & {0.900}                     & {36.86}                     & \underline {0.972}                     & 29.45                     & 0.903                     & 33.53                     & {0.939}                      & \cellcolor[HTML]{EFEFEF}32.01 & \cellcolor[HTML]{EFEFEF}0.924 \\
IDT \cite{IDT}             & 29.69                     & 0.905                     & 29.95                     &  \underline{0.898}                     & \underline{37.01}                     & {0.971}                     & 31.38                     & 0.908                     & 33.38                     & 0.937                      & \cellcolor[HTML]{EFEFEF}32.28 & \cellcolor[HTML]{EFEFEF}{0.924} \\
Semi-SwinDerain \cite{semi-derain} & 28.54                     & 0.893                     & 28.79                     & 0.861                     & 34.71                     & 0.957                     & 30.96                     & 0.909                     & 32.68                     & 0.932                      & \cellcolor[HTML]{EFEFEF}31.14 & \cellcolor[HTML]{EFEFEF}0.910 \\
LDRCNet\cite{he2023latent}    & \textbf{31.19}                     & \textbf{0.913}                     & \underline{30.60}                     & 0.892                     & 36.73                     & 0.967                     & \underline{32.89}                     &  \underline{0.917}                     & \underline{33.67}                     & \underline{0.939}                      & \cellcolor[HTML]{EFEFEF}\underline{33.02} & \cellcolor[HTML]{EFEFEF}\underline{0.926} \\
AFENet(ours)    & \underline{30.50}                     & \underline{0.918}                     & \textbf{31.22}                  & \textbf    {0.901}                 & \textbf     {37.66}                 & \textbf     {0.978}                     & \textbf{33.13}                     & \textbf{0.925}                     & \textbf{33.82}                     & \textbf{0.944}                      & \cellcolor[HTML]{EFEFEF}\textbf{33.27} & \cellcolor[HTML]{EFEFEF}\textbf{0.933} \\

\hline
\end{tabular}
\label{rain13k}
\vspace{-15pt}
\end{table*}

To effectively discern and select discriminative features that capture non-local interactions, we utilize adaptive max pooling on the input features to produce multi-scale features. These features are then concatenated using a 1 × 1 convolution coupled with GELU non-linearity. This process facilitates the integration of both local and global feature relationships. Ultimately, we enhance the output by adding the input feature back as part of residual learning, ensuring the production of clear and high-quality de-rained images.


\begin{table}[t]
\caption{Quantitative comparisons with existing state-of-the-art deraining methods in Rainds datasets.}
\begin{tabular}{ccccccc}
\hline

                              \cline{2-7} 
\multicolumn{1}{c|}{}                         & \multicolumn{2}{c|}{RS}                               & \multicolumn{2}{c|}{RD}                               & \multicolumn{2}{c}{RDS}                               \\
\multicolumn{1}{c|}{\multirow{1}{*}{method}} & \multicolumn{1}{c}{PSNR}  & \multicolumn{1}{c}{SSIM}  & \multicolumn{1}{c}{PSNR}  & \multicolumn{1}{c}{SSIM}  & \multicolumn{1}{c}{PSNR}  & \multicolumn{1}{c}{SSIM}  \\ \hline
GMM \cite{Gauss}                                 & \multicolumn{1}{c}{26.66} & \multicolumn{1}{c}{0.781} & \multicolumn{1}{c}{23.04} & \multicolumn{1}{c}{0.793} & \multicolumn{1}{c}{21.50} & \multicolumn{1}{c}{0.669} \\
JCAS \cite{gu2017joint}                                 & \multicolumn{1}{c}{26.46} & \multicolumn{1}{c}{0.786} & \multicolumn{1}{c}{23.15} & \multicolumn{1}{c}{0.811} & \multicolumn{1}{c}{20.91} & \multicolumn{1}{c}{0.671} \\
DDN \cite{DerainNet}                                & \multicolumn{1}{c}{30.41} & \multicolumn{1}{c}{0.869} & \multicolumn{1}{c}{27.92} & \multicolumn{1}{c}{0.885} & \multicolumn{1}{c}{26.85} & \multicolumn{1}{c}{0.796} \\
NLEDN \cite{li2018non}                             & \multicolumn{1}{c}{36.24} & \multicolumn{1}{c}{0.958} & \multicolumn{1}{c}{34.87} & \multicolumn{1}{c}{0.957} & \multicolumn{1}{c}{32.21} & \multicolumn{1}{c}{0.934} \\
RESCAN \cite{RESCAN}                             & \multicolumn{1}{c}{30.99} & \multicolumn{1}{c}{0.887} & \multicolumn{1}{c}{29.90} & \multicolumn{1}{c}{0.907} & \multicolumn{1}{c}{27.43} & \multicolumn{1}{c}{0.818} \\
PReNet \cite{PreNet}                            & \multicolumn{1}{c}{36.63} & \multicolumn{1}{c}{0.968} & \multicolumn{1}{c}{34.58} & \multicolumn{1}{c}{0.964} & \multicolumn{1}{c}{32.21} & \multicolumn{1}{c}{0.934} \\
UMRL \cite{URML}                                & 35.76                     & 0.962                     & 33.59                     & 0.958                     & 31.57                     & 0.929                     \\
JORDER-E \cite{JORDER}                            & 33.65                     & 0.925                     & 33.51                     & 0.944                     & 30.05                     & 0.870                     \\
Uformer \cite{uformer}                              & 40.69                     & 0.972                     & 37.08                     & 0.966                     & 34.99                     & 0.954                     \\
MSPFN \cite{MSPFN}                               & 38.61                     & 0.975                     & 36.93                     & 0.973                     & 34.08                     & 0.947                     \\

MPRNet \cite{MPRNet}                              & 40.81                     & 0.981                     & 37.03                     & 0.972                     & 34.99                     & 0.956                     \\
CCN \cite{quan2021removing}                                 & 39.17                     & 0.981                     & 37.30                     & \textbf{0.976}                     & 34.79                     & 0.957                     \\

NAFnet \cite{chen2022simple}                               & 40.39                     & 0.972                     & 37.23                     & 0.974                     & 34.99                     & 0.957                     \\
DGUNet \cite{mou2022deep}                               & {\ul 41.09}               &\textbf {0.983}               & {\ul 37.56}               &  {\ul 0.975}               & {\ul 35.34}               & {\ul 0.959}               \\
AFENet(ours)                         & \textbf{43.17}            & {\ul 0.981}            & \textbf{38.67}            &  
0.969            & \textbf{35.78}            & \textbf{0.989}            \\ \hline
\end{tabular}
\label{rds}
\end{table}

\section{Experiments}


\subsection{Implementation Details}

In our experiments, we configure the TransformerBlock with a head count of 2 and set the channel expansion factor in the GRDB to 2.66. The number of channels in the intermediate layers of our architecture is established at 128. In the training phase, we employ the Adam optimizer, starting with momentum values of 0.9 and 0.999. The initial learning rate is set at 0.0001, and we adopt a cyclic strategy for adjusting the learning rate, with a peak set at 0.0003. Our models are trained on 256 × 256 patches with batch sizes of 64 for approximately 800,000 iterations. Additionally, we implement a data augmentation strategy involving horizontal and vertical flips during training. Experiments are performed using the PyTorch framework on eight NVIDIA GeForce RTX 3090 GPUs.

\subsection{Metrics.} Following previous works \cite{JORDER}, we employ the Peak Signal-to-Noise Ratio (PSNR) \cite{PSNR} and the Structure Similarity Index Measure (SSIM) \cite{SSIM} to quantitatively assess performance on the luminance (Y) channel of synthetic datasets. For evaluating the effectiveness of our deraining approach on real-world scenes, we utilize the Naturalness Image Quality Evaluator (NIQE) \cite{NIQE} and the Blind/Referenceless Image Spatial Quality Evaluator (BRISQUE) \cite{mittal2011blind} for our assessments.. It is important to note that higher  values of PSNR and SSIM signify improved performance, whereas lower scores for NIQE and BRISQUE are desirable.

\subsection{Comparisons with State-of-the-art Methods}
To demonstrate the superiority of our method, we compare our model with twenty-four existing state-of-the-art deraining methods on two prevailing deraining benchmarks, including GMM \cite{Gauss}, DerainNet \cite{DerainNet}, JCAS \cite{gu2017joint}, DDN \cite{DerainNet}, SEMI \cite{SEMI}, DIDMDN \cite{DIDMDN}, UMRL \cite{URML}, RESCAN \cite{RESCAN}, NLEDN \cite{li2018non}, SPANet\cite{SPANet}, PReNet \cite{PreNet}, JORDER-E \cite{JORDER}, MSPFN \cite{MSPFN}, CRDNet \cite{peng}, MPRNet \cite{MPRNet}, IDLIR \cite{IDLIR}, Uformer-B \cite{uformer}, CCN \cite{quan2021removing}, IDT \cite{IDT}, Semi-SwinDerain \cite{semi-derain}, LDRCNet\cite{he2023latent}, Uformer \cite{uformer}, NAFNet \cite{chen2022simple} and DGUNet \cite{mou2022deep}.

\begin{figure*}[t]
\centerline{\includegraphics[width=1.0\linewidth]{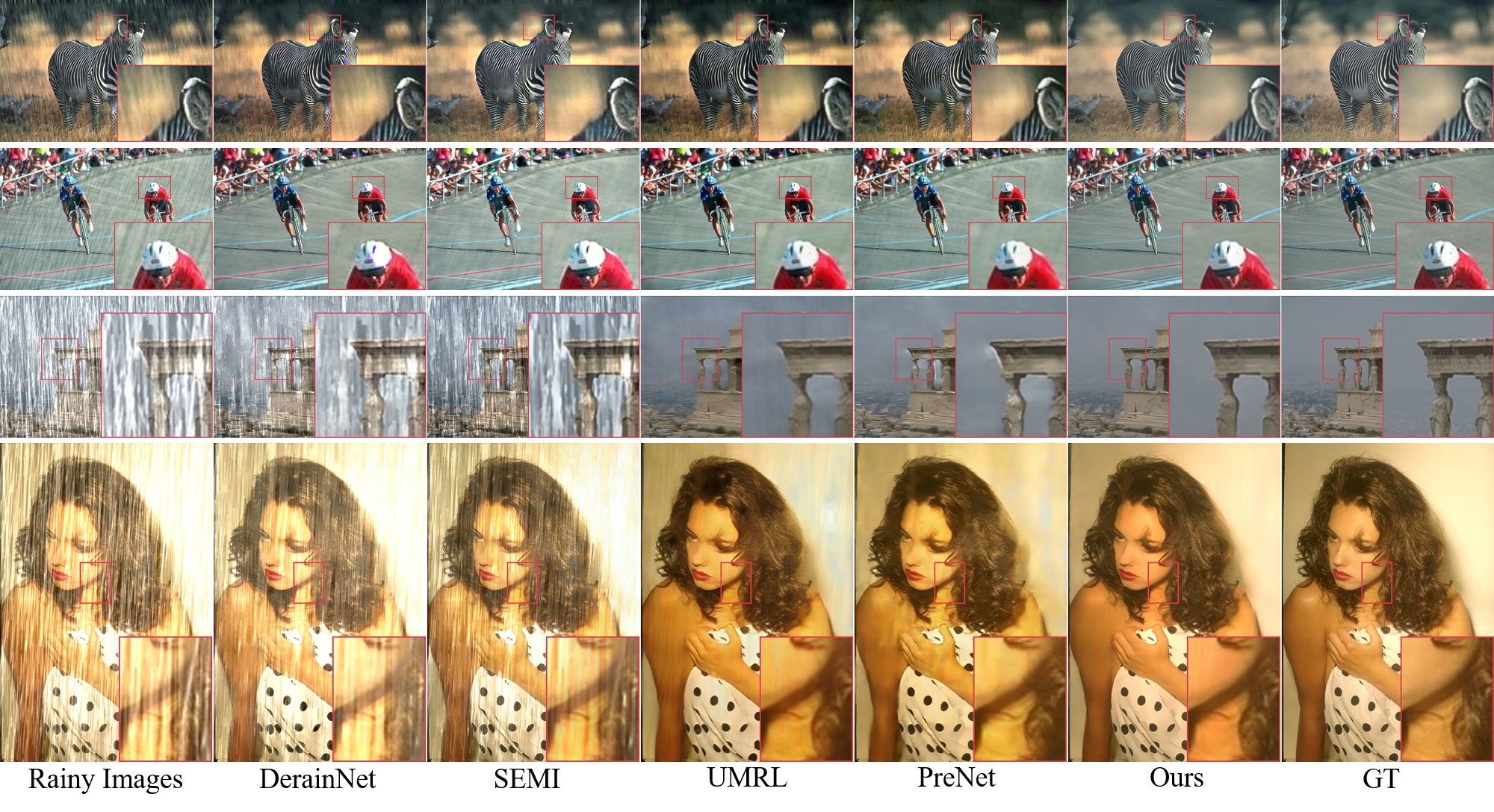}}
\caption{Visual Comparison with Existing Deraining Methods on the Rain100H and Test100 Datasets. The highlighted red boxes focus on the detailed aspects of the deraining results. Please zoom in for an enhanced view.}
\label{fig:visual}
\end{figure*}

\subsection{Datasets}
\textbf{Synthetic Data.}
Following in \cite{MPRNet}, we train our model on the widely-used Rain13K dataset, which comprises 13,712 clean-rain image pairs sourced from multiple synthetic datasets. These test sets include Rain100H \cite{JORDER}, Test100 \cite{Rain800}, Rain100L \cite{JORDER}, Test2800 \cite{DerainNet}, and Test1200 \cite{DIDMDN}. Additionally, to verify the effectiveness of our method across different rain scenes, we employ the RainDS benchmark datasets \cite{chen2023sparse} for both training and evaluation. The RainDS dataset, which includes image pairs with rain streaks (RS), raindrops (RD), and both (RDS), comprises 3600 pairs in total. Of these, 3000 are allocated for training purposes and the remaining 600 are utilized for testing.

\textbf{Real-world Data.}
To assess the efficacy of our proposed method in real-world scenarios, we use the Real15 \cite{JORDER} and Real300 \cite{Real300} datasets. These datasets, containing 15 and 300 rainy images captured from real-world rainy scenes, respectively, are employed to evaluate our methods and compare them with existing deraining approaches.

\begin{table}[]
\caption{NIQE($\downarrow$)/BRISQUE($\downarrow$) performance comparisons on the real-world datasets Real15 and Real300.}
\setlength{\tabcolsep}{0.4mm}{
\begin{tabular}{llllll}
\hline
\multicolumn{1}{c}{Datasets} & \multicolumn{1}{c}{UMRL\cite{URML}} & \multicolumn{1}{c}{PreNet\cite{PreNet}} & \multicolumn{1}{c}{MSPFN\cite{MSPFN}} & \multicolumn{1}{c}{MPRNet\cite{MPRNet}} & \multicolumn{1}{c}{AFENet(Ours)} \\ \hline
Real15                       & 16.60/24.09              & 16.04/25.29                & 17.03/23.60               & 16.48/23.92                & \textbf{15.90/23.18}            \\
Real300                      & 15.78/26.39              & 15.12/\textbf{23.57}                & 15.34/28.27               & 15.08/28.69                & \textbf{15.01}/26.34            \\ \hline
\end{tabular}
}
\label{tab:real}
\end{table}

\subsection{Quantitative Analysis}
To illustrate the superiority of our method, we conduct comparisons with twenty-four existing state-of-the-art deraining methods across two public benchmarks. The results, as presented in Table \ref{rain13k} and Table \ref{rds}, show that our method achieves the highest performance in terms of PSNR and SSIM. Notably, on the Rain13k benchmark, our method outperforms the previous state-of-the-art LDRCNet by 0.23 dB in PSNR and 0.007 in SSIM. This indicates our method's effectiveness in removing a variety of rain streaks and reconstructing more detailed imagery. Additionally, as Table \ref{rds} demonstrates, our method outperforms others on the RainDS benchmark, which includes both rain streaks and raindrops. Specifically, we exceed the performance of the leading deraining method, DGUNet \cite{mou2022deep}, by 2.08 dB in PSNR for the RS scene and by 1.11 dB in PSNR for the RD scene. This highlights our method's ability to effectively eliminate complex rainy degradation patterns and deliver superior deraining results.

We conduct comparisons with existing approaches under two publicly available real rainy scene conditions to further demonstrate the generalization capabilities of our proposed method. The results are presented in Table \ref{tab:real}. We observe that our method achieves the best performance on the no-reference metrics NIQE and BRISQUE in real-world scenarios. This demonstrates the generalizability of our approach in real-world scenarios and underscores the potential of the proposed method for deployment in practical environments.

\subsection{Qualitative Analysis}
To showcase the visual superiority of our approach, we perform comparative analyses with existing deraining methods on Rain100H and Test100, as shown in Fig. \ref{fig:visual}. It is evident that our method more effectively removes rain streaks compared to others like SEMI and UMRL, which leave residual rain streaks in their deraining results. Our method particularly excels in detail restoration. Thanks to our innovative Frequency Enhancement Module (FEM) and region-adaptive Frequency Aggregation Module (FAM), it not only removes rain streaks but also preserves reliable image details. This demonstrates our method's consistent improvement across various frequency bands, underscoring its robustness and generalization capabilities in diverse rainy conditions.

\begin{table}[h]
\caption{studies on different settings}
\setlength{\tabcolsep}{3.7mm}{
\begin{tabular}{c|ccccc}
\hline
     & S1 & S2 & S3 & S4 & \textbf{Ours} \\ \hline
PSNR & 32.48   & 32.04   & 32.98   & 32.44   & \textbf{33.27}     \\ \hline
SSIM & 0.919   & 0.912   & 0.929   & 0.917   & \textbf{0.933}     \\ \hline
\end{tabular}}
\label{ablation}
\end{table}
\subsection{Ablation Study}
\label{Ablation Study}
To validate the effectiveness of each proposed component in our proposed method, we perform the following ablation studies: (S1): Without frequency decomposition module, we simply feed a full-frequency rainy image into the branch of the frequency enhancement module; (S2): The frequency decomposition module implemented using convolution is replaced with DCT for frequency decomposition; (S3): Without interaction between different frequency branches; (S4): Using concatenation to replace the frequency aggregation module. In Table \ref{ablation}, we can observe that all components are crucial for our AFENet. For example, comparing S1, S2, and our method, which demonstrates frequency separation using convolution for frequency separation is effective. The performance of the proposed method degrades by 0.29 dB and 0.004 on PSNR and SSIM without interaction, demonstrating the superiority of our frequency interaction strategy. The performance of the proposed method degrades 0.83 dB and 0.016 on PSNR and SSIM without aggregation, demonstrating that long-range dependencies from multi-scale feature representations can be used to identify and restore valuable features for clean image reconstruction more effectively.

\section{Conclusion}
In this paper, we introduce an innovative end-to-end Adaptive Frequency Enhancement Network (AFENet), leveraging our uniquely designed Frequency Enhancement Module and Feature Aggregation Module. These modules work in tandem to accomplish feature enhancement and fusion across different frequency branches. AFENet demonstrates a remarkable ability to adaptively enhance features in various frequency components, thus effectively producing high-quality images that are free from rain artifacts. The proposed network is evaluated on four public deraining benchmarks, including both synthetic and real. Extensive experiments have consistently shown that AFENet not only excels beyond the current state-of-the-art deraining methods but also delivers visually impressive enhancement results.


\bibliographystyle{ieeebib}
\bibliography{reference}


\end{document}